\documentclass[english]{article}
\usepackage[utf8]{inputenc}
\usepackage[T1]{fontenc}
\usepackage{babel}
\usepackage{amsmath}
\usepackage{amssymb}
\usepackage{graphicx}
\usepackage{fancyhdr}
\usepackage{todonotes}
\usepackage[hyphens]{url}

\usepackage{subfig}
\usepackage{hyperref}

\begin{document}

\title{DUKweb: Diachronic word representations from the UK Web Archive corpus}

\author{Adam Tsakalidis\textsuperscript{1,2}, Pierpaolo
Basile\textsuperscript{3}\\
Marya Bazzi\textsuperscript{1,4,5}, Mihai Cucuringu\textsuperscript{1,5}, and Barbara McGillivray\textsuperscript{*1,6}
}

\maketitle
\thispagestyle{fancy}

1. The Alan Turing Institute, London, United Kingdom 2. Queen Mary University of London, London, UK 3. University of Bari, Bari, Italy 4. University of Warwick, Coventry, United Kingdom 5. University of Oxford, Oxford, United Kingdom 6. University of Cambridge, Cambridge, United Kingdom
{*}corresponding author:
\url{bm517@cam.ac.uk}
\begin{abstract}

Lexical semantic change (detecting shifts in the meaning and usage of words) is an important task for social and cultural studies as well as for Natural Language Processing applications. Diachronic word embeddings (time-sensitive vector representations of words that preserve their meaning) have become the standard resource for this task. However, given the significant computational resources needed for their generation, very few resources exist that make diachronic word embeddings available to the scientific community. 

In this paper we present DUKweb, a set of large-scale resources designed for the diachronic analysis of contemporary English.  DUKweb was created from the JISC UK Web Domain Dataset (1996-2013), a very large archive which collects resources from the Internet Archive that were hosted on domains ending in `.uk'. DUKweb consists of a series word co-occurrence matrices and two types of word embeddings for each year in the JISC UK Web Domain dataset. We show the reuse potential of DUKweb and its quality standards via a case study on word meaning change detection.
\end{abstract}

\section{Background \& Summary} 
Word embeddings, dense low-dimensional representations of words as real-number vectors \cite{mikolov}, are widely used in many Natural Language Processing (NLP) applications, such as part-of-speech tagging, information retrieval, question answering, sentiment analysis, and are employed in other research areas, including  biomedical sciences \cite{zhang} and scientometrics \cite{chinazzi}. One of the reasons for this success is that such representations allow us to perform vector calculations in geometric spaces which can be interpreted in semantic terms (i.e. in terms of the similarity in the meaning of words). This follows the so-called distributional hypothesis \cite{lenci_2008}, according to which words occurring in a given word's context contribute to some aspects of its meaning, and semantically similar words share similar contexts. In Firth's words this is summarized by the quote ``You shall know a word by the company it keeps'' \cite{firth}. 

Vector representations of words can take various forms, including count vectors, random vectors, and word embeddings. The latter are nowadays most commonly used in NLP research and are based on neural networks which transform text data into vectors of typically 50-300 dimensions. One of the most popular approaches for generating word embeddings is word2vec \cite{mikolov}. 
A common feature of such word representations is that they are labour-intensive and time-consuming to build and train. Therefore, rather than training embeddings from scratch, in NLP it is common practice to use existing pre-trained embeddings which have been made available to the community. These embeddings have typically been trained on very large web resources, for example Twitter, Common Crawl, Gigaword, and Wikipedia \cite{bojanowski2016enriching,cieliebak_etal_2017}.

Over the past few years NLP research has witnessed a surge in the number of studies on diachronic word embeddings \cite{kutuzov-etal-2018-diachronic, tahmasebi18}. 
One notable example of this emerging line of research is \cite{hamilton2016diachronic}, where the authors proposed a method for detecting semantic change using word embeddings trained on the Google Ngram corpus \cite{linetal} covering 8.5 hundred billion words from English, French, German, and Chinese historical texts. The authors have released the trained word embeddings on the project page \cite{hamilton_data}. 
The embeddings released in \cite{hamilton2016diachronic} have been successfully used in subsequent studies \cite{Garg2017,hamilton2016cultural} and over time further datasets of diachronic embeddings have been made available to the scientific community. The authors of \cite{Kim2014} released word2vec word embeddings for every 5 year-period, 
trained on the 10 million 5-grams from the English fiction portion of the Google Ngram corpus \cite{google_data}. The authors of \cite{Novel2018} have released different versions of word2vec embeddings trained on the Eighteenth-Century Collections Online (ECCO-TCP corpus), covering the years 1700-1799 \cite{heuser_data}. These include embeddings trained on five twenty-year periods for 150 million words randomly sampled from the ``Literature and Language'' section of this corpus. Another set of diachronic word embeddings was released as part of a system for diachronic semantic search on text corpora based on the Google Books Ngram Corpus (English Fiction and German subcorpus), the Corpus of Historical American English, the Deutsches Textarchiv `German Text Archive' (a corpus of ca. 1600-1900 German), and the Royal Society Corpus (containing the first two centuries of the Philosophical Transactions of the Royal Society of London) \cite{Hellrich2017}.\footnote{\url{http://jeseme.org/help.html}, last accessed 27/11/2020.}

The only example of trained word diachronic embeddings covering a short and recent time period are available in \cite{shoemark2019room_data} and were built following the methodology described in \cite{shoemark2019room}. The authors trained monthly word embeddings from the tweets available via the Twitter Streaming API from 2012 to 2018 and comprising over 20 billion word tokens.

The word embeddings datasets surveyed in this section are useful resources for researchers conducting linguistic diachronic analyses or developing NLP tools that require data with a chronological depth. However, more steps are needed in order to process these embeddings further.

We present DUKweb, a rich dataset comprising diachronic embeddings, co-occurrence matrices, and time series data which can be directly used for a range of diachronic linguistic analysis aimed an investigating different aspects of recent language change in English. DUKweb was created from the JISC UK Web Domain Dataset (1996-2013), a very large archive which collects resources from the Internet Archive hosted on domains ending in ‘.uk’.
DUKweb consists of three main components: 
\begin{enumerate}
    \item co-occurrences matrices for each year built by relying on the original text extracted from the JISC UK Web Domain Dataset;
    \item a set of different word embeddings (Temporal Random Indexing and word2vec) for each year;
    \item time series of words' change in representation across time.
\end{enumerate}
DUKweb can be used for several time-independent NLP tasks, including word similarity, relatedness, analogy, but also for temporally dependent tasks, such as semantic change detection (i.e., tracking change in word meaning over time), a task which has received significant attention in recent years \cite{schlechtweg-etal-2020-semeval,diacrita_evalita2020}.

The main innovative features of our contribution are:
\begin{itemize}
    \item Variety of object types: we release the full package of data needed for diachronic linguistic analyses of word meaning: co-occurrence matrices, word embeddings, and time series.
    \item Size: we release word vectors trained on a very large contemporary English diachronic corpus of 1,316 billion word occurrences spanning the years 1996-2013; the total size of the dataset is 330GB;
    \item Time dimension: the word vectors have been trained on yearly portions of the UK web archive corpus corpus, which makes them ideally suited for many diachronic linguistic analyses;
    \item Embedding types: we release count-based and prediction-based types of word embeddings: Temporal Random Indexing vectors and word2vec vectors, respectively, and provide the first systematic comparison between the two approaches;
\end{itemize}

None of the other existing datasets offer researchers all the above features. The surveyed datasets are based on corpora smaller than the UK Web archive JISC dataset, the biggest one being 850 billion words vs. 1316 billion words \cite{linetal}. Moreover, only the Twitter embedding resource \cite{shoemark2019room_data} was specifically built to model recent language change (2012-2018). On the other hand, recent workshops on semantic change detection \cite{diacrita_evalita2020,schlechtweg-etal-2020-semeval} study the semantic change across the two distinct time periods and therefore lack the longitudinal aspect of our resources. In addition to being based on a much larger corpus with a longitudinal component, DUKweb can be readily used to study semantic change in English between 1996 and 2013 and therefore to investigate the effects of various phenomena such as the expansion of the World Wide Web or social media on the English language. Finally, DUKweb offers researchers a variety of object types and embedding types; as we show in our experiments, these capture the semantic change of different words and can therefore be leveraged in conjunction for further analysis in future work.

\section{Methods}\label{sec:method}

\begin{figure}[htb]
	\centering
	\includegraphics[width=12cm]{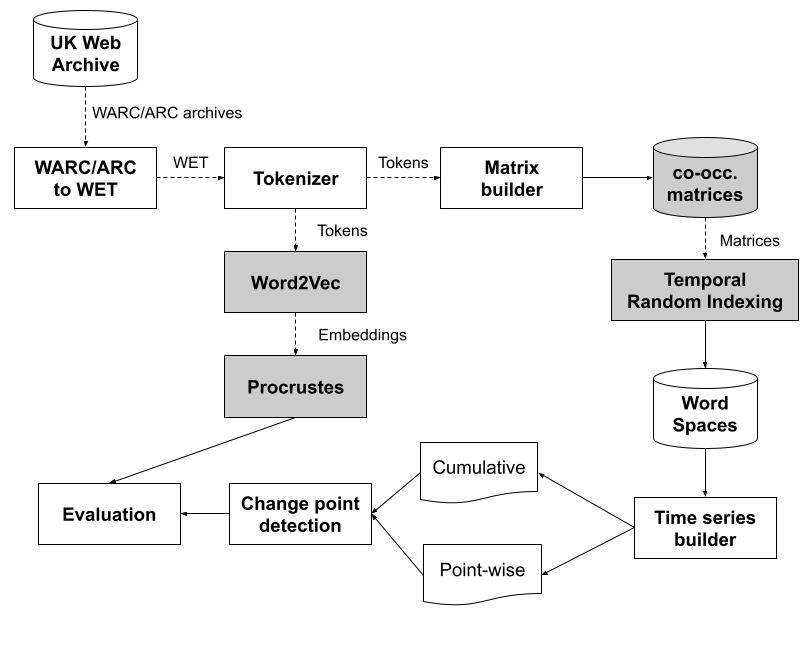}
	\caption{Flowchart of the creation of DUKweb.}\label{fig:dataFlowCorpus}
\end{figure}

The flow chart in Figure \ref{fig:dataFlowCorpus} illustrates the process from corpus creation to the construction of DUKweb. In the next sub-sections we provide details about how the three parts of the dataset were built.

\subsection{Source data}\label{subsec:sourceData}

We used the JISC UK Web Domain Dataset (1996-2013) \cite{jisc}, which collects resources from the Internet Archive (IA) that were hosted on domains ending in ‘.uk’, and those that are required in order to render ‘.uk‘ pages.
The JISC dataset contains resources crawled by the IA Web Group for different archiving partners, the Web Wide crawls and other miscellaneous crawls run by IA, as well as data donations from Alexa (\url{https://www.alexa.com}) and other companies or institutions, therefore we do not have access to all the crawling configurations used by the different partners. The dataset contains not only HTML pages and textual resources, but also video, images and other types of files.

The JISC dataset is composed of two parts: the first part contains resources from 1996 to 2010 for a total size of 32TB; the second part contains resources from 2011-2013 for a total size of 30TB. The JISC dataset cannot be made generally available, but can be used to generate derived datasets like DUKweb, which can be released publicly for research use. 

\subsection{Text extraction and pre-processing}\label{subsec:processing}

The first step in the creation of DUKweb consisted in processing the JISC web archive in order to extract its textual resources. For this purpose, we extracted the text from resources such as TXT files and parsed HTML pages. We used the jsoup library (\url{https://jsoup.org/}) for parsing HTML pages.
The original JISC dataset contains files in the ARC and WARC formats, standard formats used by IA for storing data crawled from the web as sequences of content blocks. The WARC format\footnote{\url{https://commoncrawl.org/2014/04/navigating-the-warc-file-format/}, last accessed 27/11/2020} is an enhancement of the ARC format supporting a range of features including metadata and duplicate events. 
We converted the ARC and WARC archives into the WET format, a standard format for storing plain text extracted from ARC/WARC archives. The output of this process provided about 5.5TB of compressed WET archives.

The second step consisted in tokenizing the WET archives. For this purpose, we used the StandardAnalyzer\footnote{\url{https://lucene.apache.org/core/7_3_1/core/index.html}, last accessed 27/11/2020.} provided by the Apache Lucene 4.10.4 API.\footnote{\url{https://lucene.apache.org/core/}, last accessed 27/11/2020.} This analyzer also provides a standard list of English stop words. The size of the tokenized corpus is approximately 3TB, with a vocabulary size of 29 million tokens and about 1200 billion occurrences.
We did not apply any further text processing steps such us lemmatization or stemming because our aim was to build a language independent pipeline.

\subsection{Co-occurrence matrices}

The co-occurrence matrices of DUKweb store co-occurrence information for each word token in the JISC dataset processed as described in section \ref{subsec:processing}. For the construction of co-occurrence matrices, we focused on the 1,000,000 most frequent words.

In order to track temporal information, we built a co-occurrence matrix for each year from  1996 to 2013.
Each matrix is stored in a compressed text format, with one row for each token, where each row contains the token and the list of tokens co-occurring with it. Following standard practice in NLP, we extracted co-occurrence counts by taking into account a window of five words to the left and five words to the right of the target word \cite{levy2014dependency,zhao2017ngram2vec,tsakalidis2019}.

\subsection{Word Embeddings}

We constructed semantic representations of the words occurring in the processed JISC dataset by training word embeddings for each year using two approaches: Temporal Random Indexing  and the word2vec algorithm (i.e., skip-gram with negative sampling). The next subsections provide details of each approach.

\subsubsection{Temporal Random Indexing} \label{sec:tri}

The first set of word embeddings of DUKweb was trained using Temporal Random Indexing (TRI) \cite{clic2014a,ref88,basile-mcgillivray_2018}. We further developed the TRI approach in three directions: 1) we improved the system to make it possible to process very large datasets like the JISC UK Web Domain Dataset; 2) we introduced a new way to weigh terms in order to reduce the impact of very frequent tokens; 3) compared to our previous work on the same topic \cite{basile-mcgillivray_2018}, we proposed a new ``cumulative'' approach for building word vectors.

The idea behind TRI is to build different word spaces for each time period under investigation. 
The peculiarity of TRI is that word vectors over different time periods are directly comparable because they are built using the same random vectors.
TRI works as follows:
\begin{enumerate}
	\item Given a corpus $C$ of documents and a vocabulary $V$ of terms\footnote{$V$ contains the terms that we want to analyse, typically, the top $n$ frequent terms.} extracted from $C$, the method assigns a random vector $r_i$ to each term $t_i \in V$. A random vector is a vector that has values in the set \{-1, 0, 1\} and is sparse, with few non-zero elements randomly distributed along its dimensions. The sets of random vectors assigned to all terms in $V$ are near-orthogonal;
	\item The corpus $C$ is split into different time periods $T_k$ using temporal information, for example the year of publication; 
	\item For each period $T_k$, a word space $WS_k$ is built. Each of the terms of $V$ occurring in $T_k$ is represented by a semantic vector. The semantic vector $sv_i^{k}$ for the $i$-th term in $T_k$ is built as the sum of all the random vectors of the terms co-occurring with $t_i$ in $T_k$.
	Unlike the approach proposed in \cite{basile-mcgillivray_2018}, the $sv_i^{k}$ is not initialized as a zero vector, but as the the semantic vectors $sv_i^{k-1}$  built in the previous period. Using this approach we are able to collect semantic features of the term across time. If the $sv_i^{k-1}$ is not available,\footnote{The $sv_i^{k-1}$ is not available when the term $t_i$ appears for the first time in $T_k$.} the zero vector is used.
	When computing the sum, we apply some weighting to the random vector. To reduce the impact of very frequent terms, we use the weights $\sqrt{\frac{th \times C_k}{\#t_i^k}}$, where $C_k$ is the total number of occurrences in $T_k$ and $\#t_i^k$ is the number of occurrences of the term $t_i$ in $T_k$. The parameter $th$ is set to 0.001.
\end{enumerate}

This way, the semantic vectors across all time periods are comparable since they are the sum of the same random vectors.

\paragraph{Time series}\label{sec:time-series}

For each term $t_i$ DUKweb also contains a time series $\Gamma(t_i)$, which can be used to track a word's meaning change over time. The time series are sequences of values, one value for each time period, and represent the semantic shift of that term in the given period. We adopt several strategies for building the time series. 
The baseline approach is based on term log-frequency, where each value in the series is defined as $\Gamma_k(t_i) = \log \left(\frac{\#t_i^k}{C_k}\right)$. 

In addition to the baseline, we devised two other strategies for building the time series: 
\begin{description}
	\item[point-wise:] $\Gamma_k(t_i)$ is defined as the cosine similarity between the semantic vector of $t_i$ in the time period $k$, $sv_i^{k}$, and the semantic vector of $t_i$ in the previous time period, $sv_i^{k-1}$. This way, we capture semantic change between two time periods;
	\item[cumulative:] we build a cumulative vector $sv_i^{C_{k-1}}=\sum_{j=0}^{k-1}sv_i^{j}$ and compute the cosine similarity of this cumulative vector and the vector $sv_i^{k}$. The idea behind this approach is that the semantics of a word at point $k-1$ depends on the semantics of the word in all the previous time periods. The cumulative vector is the vector sum of all the previous word vectors \cite{Mitchell2010}.
\end{description}

\subsubsection{Skip-gram with Negative Sampling} \label{sec:word2vec}

The second approach we followed for generating word representations is based on the of skip-gram with negative sampling (SGNS) algorithm \cite{mikolov2013}. The skip-gram model is a two-layer neural network that aims at predicting the words (context) surrounding a particular word in a sentence. The training process is performed on a large corpus, where samples of \{context, word\} pairs are drawn by sliding a window of $N$ words at a time. The resulting word vectors can appropriately represent each word based on the context in which it appears so that the distance between similar words is small and the analogy of word pairs like  (\emph{king}, \emph{queen}) and (\emph{uncle}, \emph{aunt}) is maintained. Over the past few years SGNS has been widely employed and its efficiency has been demonstrated in several studies on semantic change  \cite{hamilton2016diachronic,schlechtweg2019,tsakalidis2019, schlechtweg-etal-2020-semeval}.

We split the pre-processed corpus into yearly bins, as in the case of TRI, and train one language model per year.\footnote{We refrain from using the years 1996-1999 for SGNS, due to their considerably smaller size compared to the rest of the years in our processed collection, which could result into noisy word representations.} Our skip-gram model then learns a single 100-dimensional representation for each word that is present at least 1,000 times on each year independently, i.e. 3,910,329 words. We used the implementation of skip-gram with negative sampling as provided in gensim\footnote{\url{https://radimrehurek.com/gensim/}}, using 5 words as our window size and training for 5 epochs for each year while keeping the rest of hyperparameters on their default values.
\footnote{In previous work \cite{tsakalidis2019}, we selected the 47.8K words occurring in all years in both our corpus and in the entry list of the Oxford English Dictionary. Importantly, very common words (e.g., ``facebook'') that appeared only after a certain point in time are not included in our previous analysis.}

\paragraph{Orthogonal Procrustes} 
In contrast to TRI, a drawback of training independent (i.e., one per year) SGNS models is the fact that the resulting word vectors are not necessarily aligned to the same coordinate axes across different years~\cite{hamilton2016diachronic}. 
In particular, SGNS models may result in arbitrary linear transformations, which do not affect pairwise cosine-similarities within-years, but prevent meaningful comparison across years.

To align the semantic spaces, we follow the Procrustes analysis 
from \cite{hamilton2016diachronic}. Denote by $W^{(t)}\in \mathbb{R}^{n\times m}$ the matrix of word embeddings in year $t$. The orthogonal Procrustes problem consists of solving 
\begin{equation}
    \min_{Q} \vert\vert W^{(t)}Q - W^{(t+1)}\vert\vert_F, \quad\text{subject to } Q^TQ = I\,,
    \label{OPP}
\end{equation}
where  $Q \in\mathbb{R}^{m\times m}$ is the rotation matrix and $I$ is the ${m\times m}$ identity matrix. We align the word embedding in year $t$ to their respective embeddings in year $t+1$ by finding a translation, rotation, and scaling of $W^{(t)}$ that minimizes its distance to $W^{(t+1)}$ as measured by the Frobenius norm. The optimisation problem in~\eqref{OPP} can be solved using the singular value decomposition of $W^{(t)}(W^{(t+1)})^T$. We can then use the cosine distance between the vector representation of a word across aligned embeddings as a measure of the semantic displacement of that word. 

An alternative approach would be to initialise the embeddings of the year $t+1$ with the resulting representations of the year $t$ \cite{Kim2014}. However, this would demand sequential -- and thus longer -- processing of our data and it is not clear whether the resulting representations capture the semantics of the words more effectively, as demonstrated in recent work on semantic change detection \cite{shoemark2019room}. 
Another potential approach to consider stems from the generalized orthogonal Procrustes problem which simultaneously considers all the available embeddings at once and aims to optimally identify an \textit{average embedding}  which is simultaneously close to all the input embeddings \cite{pumir2019generalized}. This contrasts with our approach where only pairwise Procrustes alignments are performed. 

\paragraph{Time Series}
We construct the time series of a word's similarity with itself over time, by measuring the cosine distance of its aligned representation in a certain year (2001-2013) from its representation in the year 2000. Recent work \cite{shoemark2019room} has demonstrated that the year used for reference matters (i.e., $W^{(t)}$ in Eq. 1). We intuitively opted to use the year 2000 as our referencing point, since it is the beginning of our time period under examination. In order to construct time series in a consistent manner, we only use the representations of the words that are present in each year.

\section{Data Records}
This section describes each data record associated with our dataset, which  is available on the British Library repository (\url{ https://doi.org/10.23636/1209}). 

\subsection{Co-occurrences matrices}
\begin{figure}[htb]
	\centering
	\includegraphics[width=10cm]{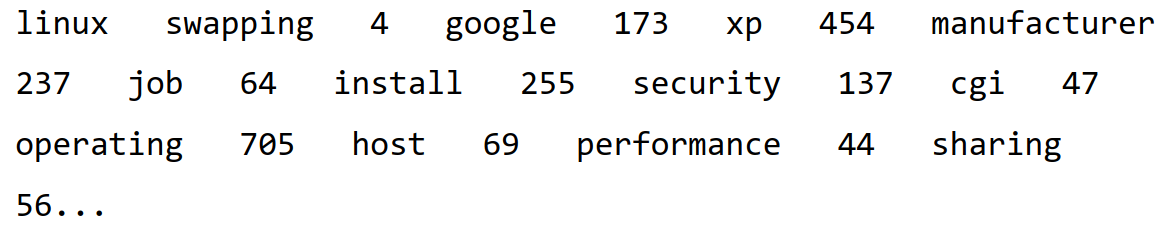}
	\caption{Example of co-occurrence matrix for the word \textit{linux} (year 2011) in DUKweb.}\label{fig:coocc}
\end{figure}

The first part of our dataset consists of the co-occurrences matrices.
We built a co-occurrence matrix from the pre-processed JISC dataset for each year from 1996 to 2013.
Each matrix is stored in a compressed text format, with one row per token. Each row reports a token and the list of tokens co-occurring with it. An example for the word \textit{linux} is reported in Figure \ref{fig:coocc}, which shows that the token \textit{swapping} co-occurs 4 times with \textit{linux}, the word \textit{google} 173 times, and so on.

Table \ref{tab:statOcc} reports the size of the vocabulary and the associated compressed file size, for each year. The total number of tokens considering only the terms in our vocabulary is 1,316 billion. 

\begin{table}[htb]
\centering
\begin{tabular}{l|r|r|r|}
Year & Vocabulary Size  & \#co-occurrences & File Size \\
\hline
1996 & 454,751    & 1,201,630,516 & 645.6MB  \\
1997 & 711,007    & 17,244,958,174 & 2.7GB   \\
1998 & 704,453    & 10,963,699,018 & 2.4GB   \\
1999 & 769,824    & 32,760,590,881 & 3.6GB   \\
2000 & 847,318    & 107,529,345,578 & 5.8GB  \\
2001 & 911,499    & 197,833,301,500 & 9.2GB \\
2002 & 945,565    & 274,741,483,798 & 11GB  \\
2003 & 992,192    & 539,189,466,798  & 14GB \\
2004 & 1,040,470  & 975,622,607,090 & 18.2GB \\
2005 & 1,060,117  & 793,029,668,228 & 16.9GB  \\
2006 & 1,076,523  & 721,537,927,839 & 16.7GB  \\
2007 & 1,093,980  & 834,261,488,677 & 18.1GB \\
2008 & 1,105,511  & 1,067,076,347,615 & 19.6GB \\
2009 & 1,105,901  & 481,567,239,481 & 14.15GB   \\
2010 & 1,125,201  & 778,111,567,761 & 16.7GB  \\
2011 & 1,145,990  & 1,092,441,542,978 & 18.9GB \\
2012 & 1,144,764  & 1,741,038,554,999 & 20.6GB \\
2013 & 1,044,436  & 393,672,000,378 & 8.9GB       
\end{tabular}
\caption{Statistics about the co-occurrences matrices in DUKweb. The first column shows the year, the second column contains the size of the vocabulary for that year in terms of number of word types, the third column contains the total number of co-occurrences of vocabulary terms for that year, and the last column shows the size (compressed) of the co-occurrence matrix file.}\label{tab:statOcc}
\end{table}

\subsection{Word embeddings}

\begin{figure}
    \centering
    \includegraphics[width=0.9\linewidth]{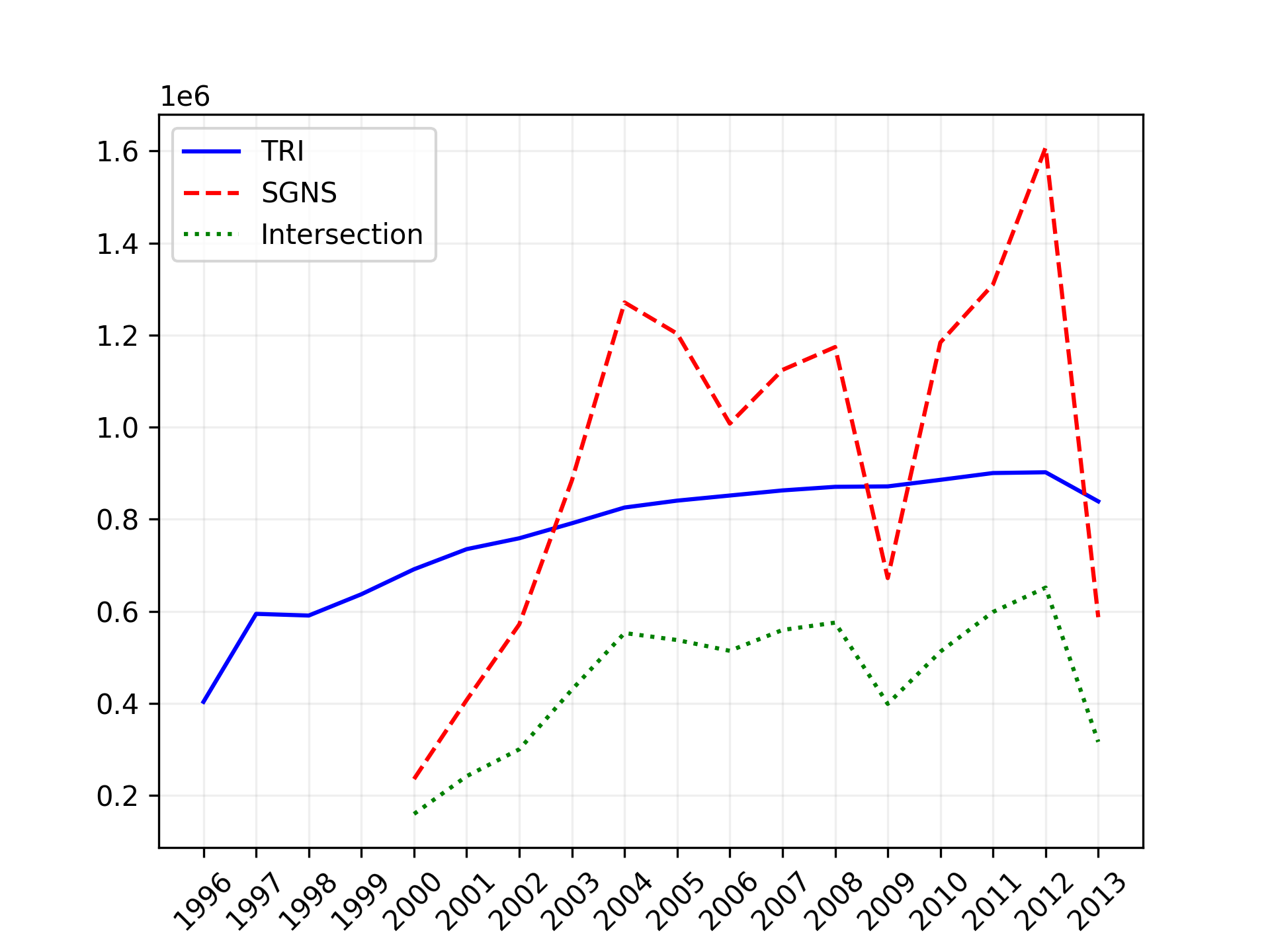}
    \caption{Number of words included in the TRI and SGNS representations contained in DUKweb, along with the size of their intersected vocabulary size, per year.}
    \label{fig:vocabintersected}
\end{figure}

The second part of the dataset contains word embeddings built using TRI and word2vec.
Both embeddings are provided in the GZIP compressed textual format, with one file for each year.
Each file stores a word embedding for each line, the line starts with a word followed by the corresponding embedding vector entries separated by spaces, for example:
\begin{verbatim}
dog 0.0019510963 -0.033144157 0.033734962...
\end{verbatim}

Table \ref{tab:statVec} shows the (compressed) file size  of each vector space. For TRI, the number of vectors (terms) for each year is equal to the vocabulary size of co-occurrences matrices as reported in Table \ref{tab:statOcc}. The TRI vector dimension is equal to 1,000, while the number of no-zero elements in random vector is set to 10. Finally, the parameter $th$ is set to 0.001, and TRI vectors are built by using the code reported in section ``Code Availability''. 
For SGNS, the total number of words represented in any year is 3,910,329. Finally, the chart in Figure~\ref{fig:vocabintersected} shows the intersected vocabulary between the two methods. The number of total terms contained in the intersection is 47,871.
We also release a version of TRI embeddings built by taking into account only the terms contained in the intersection.
In order to perform a fair evaluation, in our experiments we only take into account  the intersected vocabulary.

\begin{table}[htb]
\centering
\begin{tabular}{l|r|r|r|r|r|}
 &\multicolumn{2}{c|}{\textbf{TRI}}&\multicolumn{2}{|c|}{\textbf{SGNS}}\\\hline
Year & Vocabulary Size  & File Size   & Vocabulary Size & File Size  \\
\hline
1996 &  454,751   & 284.9MB   & -- & --\\
1997 &  711,007   &  904.8MB   & -- & --\\
1998 &  704,453   & 823.3MB    & -- &--\\
1999 &  769,824  &  1.1GB    & -- & --\\
2000 &  847,318  &  1.5GB    & 235,428&114.7MB\\
2001 &  911,499  & 1.9GB  & 407,074 & 198.2MB\\
2002 &  945,565  & 2.1GB & 571,419 & 277.8MB\\
2003 &   992,192  & 2.4GB & 884,393 & 430.4MB\\
2004 & 1,040,470 & 2.7GB & 1,270,804 & 619.1MB\\
2005 &  1,060,117 &  2.7GB & 1,202,899 & 585.4MB\\
2006 & 1,076,523  & 2.7GB  & 1,007,582 & 490.6MB\\
2007 &  1,093,980 & 2.8GB  & 1,124,179 & 548.2MB\\
2008 & 1,105,511 & 2.9GB & 1,173,870 & 572.2MB\\
2009 & 1,105,901 &  2.6GB & 671,940 & 327.6MB\\
2010 & 1,125,201  & 2.8GB & 1,183,907 & 576.8MB\\
2011 & 1,145,990  & 2.9GB & 1,309,804 & 637.7MB\\
2012 & 1,144,764  & 3.0GB  & 1,607,272 & 784.0MB\\
2013 & 1,044,436 & 1.9GB   & 587,035 & 285.6MB
\end{tabular}
\caption{Statistics about the vocabulary in terms of overall number of words and (compressed) file size, per year and per method (TRI, SGNS).}\label{tab:statVec}
\end{table}

\begin{figure}%
    \centering
    \subfloat[TRI]{{\includegraphics[width=0.47\textwidth, trim=1.2cm 0cm 1.5cm  0cm,clip]{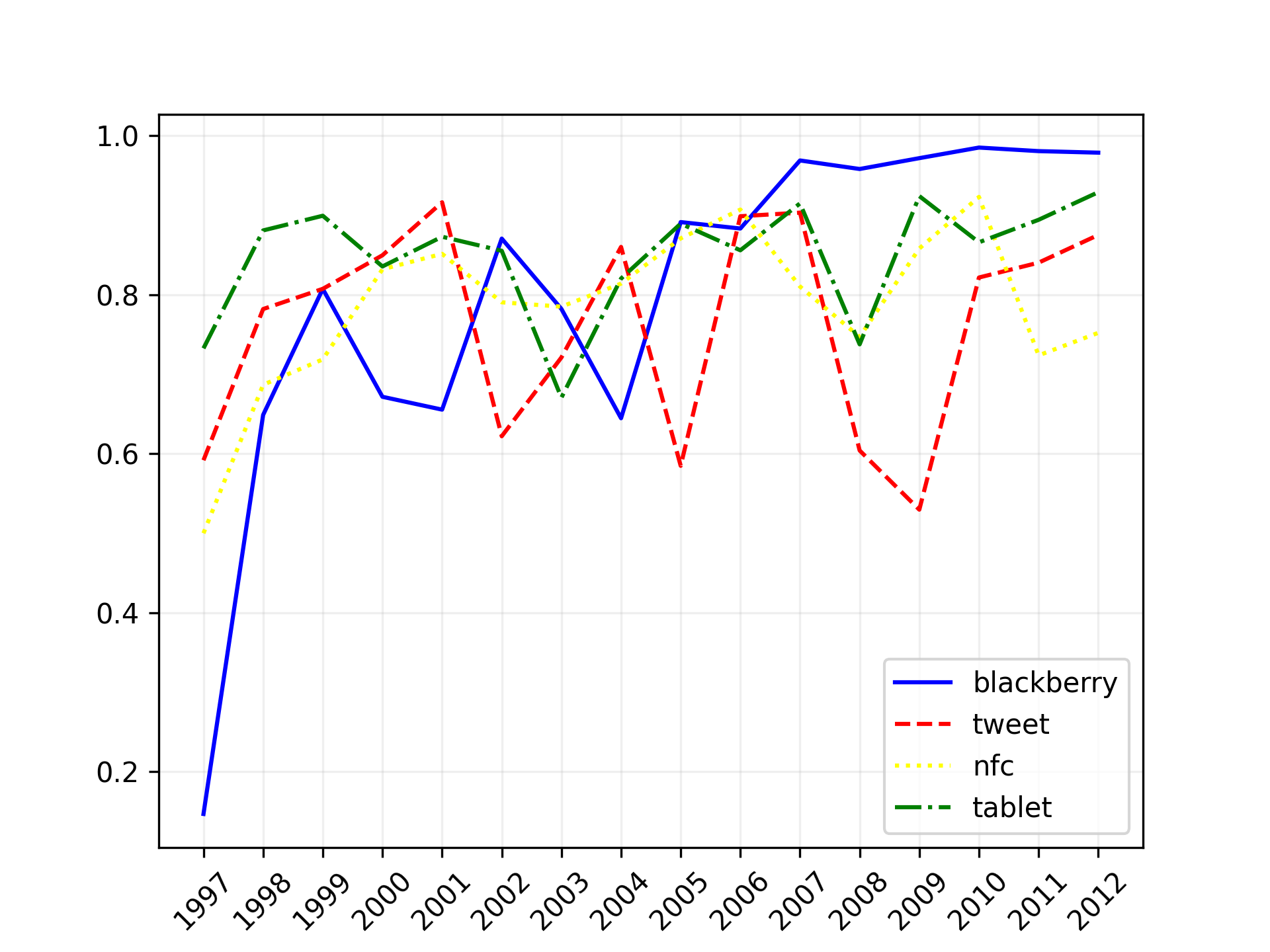} }}%
    \quad
    \subfloat[SGNS]{{\includegraphics[width=0.47\textwidth, trim=1.2cm 0cm 1.5cm  0cm,clip]{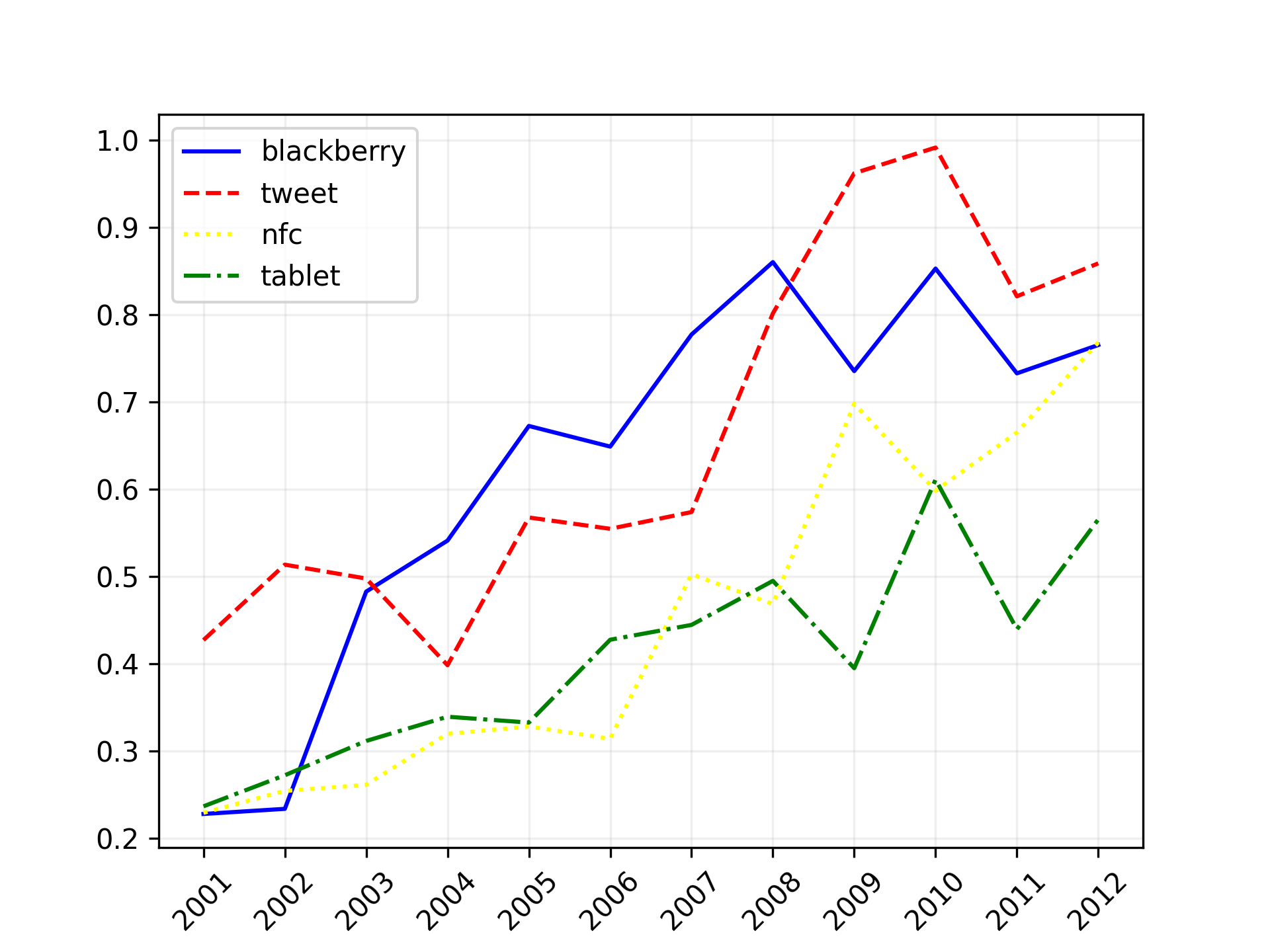} }}
    \caption{Time series based on TRI and SGNS of four words whose semantics have changed between 2001-2013 according to the Oxford English Dictionary (i.e., they have acquired a new meaning during this time period).}%
    \label{fig:plotBlackberry}%
\end{figure}

\subsection{Time series}
The last part of the dataset contains a set of time series in CSV format computed using the different strategies described in Section~\ref{sec:time-series}. 
For each time series we extract a list of change points (see Section~\ref{sec:time-series}). The chart on Figure~\ref{fig:plotBlackberry} (a) shows the time series for different words that have acquired a new meaning after the year 2000, according to the Oxford English Dictionary. Similarly, Figure~\ref{fig:plotBlackberry} (b) shows the time series of the cosine distances of the same words that result after Orthogonal Procrustes is applied on SGNS as described in Section~\ref{sec:word2vec}. In particular, we first align the embeddings matrices of a year $T$ with respect to the year 2000 and then we measure the cosine distance of the aligned word vectors. As a result, this part of our dataset consists of 168,362 tokens and their respective differences from the year 2000 during the years 2001-2013. We further accompany these with the time series consisting of the number of  occurrences of each word in every year.

Figure~\ref{fig:plotBlackberry} shows that the semantic similarity of the word ``blackberry'' decreases dramatically in 2004, which corresponds to the change point year detected by the semantic change detection algorithm. On the other hand, Figure~\ref{fig:plotBlackberry} (b) shows that the four words are moving away with time from their semantic representation in the year 2000. We also find examples of cases of semantic shifts corresponding to real-world events in SGNS representations: for example, the meaning of the word ``tweet'' shifted more rapidly after Twitter's foundation (2006), whereas the lexical semantics of the word ``tablet'' mostly shifted in the year 2010, at the time when the first mass-market tablet -- iPad -- was released.

Furthermore, as we showed in our previous work \cite{basile-mcgillivray_2018}, it is possible to analyze the neighborhood of a word (e.g. ``blackberry'') in different years (e.g. in 2003 and 2004) in order to understand the semantic change it underwent. The list of neighborhoods can be computed as the most similar vectors in a given word space. Similarly, in the case of SGNS we can measure the cosine similarity between the word representations in different years.

\subsection{Dataset Summary}
Overall, our dataset consists of the following files:

\begin{itemize}
    \item \textbf{D-\texttt{YEAR}\_merge\_occ.gz}: co-occurrences matrix for each year.
    \item \textbf{\texttt{YEAR}.csv.zip}: The SGNS word embeddings during the year \texttt{YEAR}, as described in section \ref{sec:word2vec}.
    There are 14 files, one for each year between 2000 and 2013, and the size of each file is shown in Table~\ref{tab:statVec}.
    \item \textbf{D-\texttt{YEAR}\_merge.vectors.txt.gz}: TRI embeddings for each year in textual format. 
    \item \textbf{D-\texttt{YEAR}\_merge.vectors}: TRI embeddings for each year in binary format. This vectors can be directly used with the TRI tool.\footnote{\url{https://github.com/pippokill/tri}}
    \item \textbf{timeseries}: four files containing the timeseries for the words based on (a) their counts, as extracted from SGNS (count\_timeseries.tsv), (b) the cosine distances from their representations in the year 2000 based on SGNS (w2v\_dist\_timeseries.tsv) and (c) the time series generated via TRI-based pointwise and cumulative approaches (ukwac\_all\_point.csv, ukwac\_all\_cum.csv respectively).
    \item \textbf{vectors.elemental}: TRI embeddings for the whole vocabulary in binary format.
\end{itemize}

\section{Technical Validation}
We perform two sets of experiments in order to assess the quality of the embeddings generated via TRI and SGNS. In the first set, our goal is to measure the embeddings' ability to capture semantic properties of  words, i.e. analogies, similarities and relatedness levels. In the second set, we aim at exploring to what extent the two types of contextual vectors capture the change in meaning of words.

\subsection{Static Tasks: Word-level Semantics}
In this set of tasks we examine the ability of the word vectors to capture the semantics of the words associated to them. We work on three different sub-tasks: (a) word analogy, (b) word similarity and (c) word relatedness detection.

\subsubsection{Tasks Description} 

\paragraph{Word Analogy} \textit{Word analogy detection} is the task of identifying relationships between words in the form of ``$w_a$ is to $w_b$ as $w_c$ is to $w_d$'', where $w_i$ is a word. In our experiments, we make use of the widely employed dataset which was created by Mikolov and colleagues \cite{mikolov2013} and which contains these relationships in different categories:\footnote{We list four categories, with one example for each of them.}

\begin{itemize}
    \item Geography, e.g. \textit{Paris is to France as Madrid is to Spain.}
    \item Currencies, e.g. \textit{Peso is to Argentina as euro is to Europe.}
    \item Family, e.g. \textit{Boy is to girl as policeman is to policewoman.}
    \item Grammar rules, e.g. \textit{Amazing is to amazingly as apparent is to apparently.}
\end{itemize}
Given the vectors [$v_a$,  $v_b$,  $v_c$,  $v_d$] of the words [$w_a$,  $w_b$,  $w_c$,  $w_d$] respectively, and assuming an analogous relationship between the pairs [$w_a$, $w_b$] and [$w_c$, $w_d$] in the form described above, previous work has demonstrated that in SGNS-based embeddings, $v_a-v_b \approx v_c-v_d$. We perform this task by measuring the cosine similarity  $s_c$ that results when comparing the vector $v_c$ of the word $w_c$ against the vector resulting from $v_a-v_b+v_d$. We apply this method to all the words in the examples of the word analogy task set and we average across all examples. Higher average cosine similarity scores indicate a better model.

TRI-based embeddings are not suitable for this kind of task due to the nature of the vector space. TRI is not able to learn the linear dependency between vectors in the space, which may be due to the Random Projection that preserves all distance/similarity measures based on L2-norm, but it distorts the original space and does not preserve the original position of vectors.
We try to simulate analogy by using vector orthogonalization as the negation operator \cite{widdows2003}. In particular, the vector sum $v_b + v_d$ is orthogonalized with respect to the vector $v_a$. For each word vector the cosine similarity is computed against the vector obtained as the result of the orthogonalization, then the word with the highest similarity is selected.

We perform experiments using our TRI and SNGS word vectors. For comparison purposes, we also employ the word2vec pre-trained embeddings generated in \cite{mikolov2013} as well as the GloVe embeddings released in \cite{pennington2014}. These are well-established sources of word vectors that have been trained on massive corpora of textual data and have been employed in an extremely large body of research work across multiple NLP tasks. In particular, for word2vec we employ the 300-dimensional word vectors that were trained on Google news ($pre_{w2v}$) \cite{mikolov2013}, whereas for GloVe we use the 100-dimensional vectors trained on Twitter ($pre_{glv}$) \cite{pennington2014}. As opposed to SGNS and TRI, $pre_{w2v}$ and $pre_{glv}$ are temporally independent (i.e., there is one representation of each word through time). To allow for a fair comparison, the analysis described here and in the following sections is on the intersected vocabulary across all years, i.e. on the set of words occurring in all years in our corpus.

\paragraph{Word Similarity} In the \textit{word similarity} task, we are interested in detecting the similarity between two words. We employ the dataset in \cite{agirre2009}, which contains examples of 203 word pairs, along with their similarity score, as provided by two annotators. We only use the 187 word pairs that consist of words present in the intersected vocabulary, as before. We deploy the following experimental setting: given the word vectors $w_a$ and $w_b$ of a pair of words, we measure their cosine  similarity $sim(w_a, w_b)$; subsequently, we measure the correlation between $sim(w_a, w_b)$ and the ground truth. Higher values indicate a better model. We compare performance against the baselines $pre_{w2v}$ and $pre_{glv}$.

\paragraph{Word Relatedness} Our final experiment in this section involves the task of identifying the level of relatedness between two words. For example, the word ``computer'' is much more closely related to the word ``software'' than to the word ``news''. We employ the word relatedness dataset introduced by \cite{agirre2009}, which contains 252 examples of word pairs, along with their associated relatedness score. As in the previous tasks, we only use the 232 examples that are present in our intersected vocabulary; we deploy the same experimental setting as in the case of the Word Similarity task and compare against the two previously introduced baselines ($pre_{w2v}$, $pre_{glv}$).

\subsubsection{Results} 

\begin{figure}%
    \centering
    \hspace{-1.82cm}
    \subfloat[\centering Family]{{\includegraphics[width=0.43\textwidth]{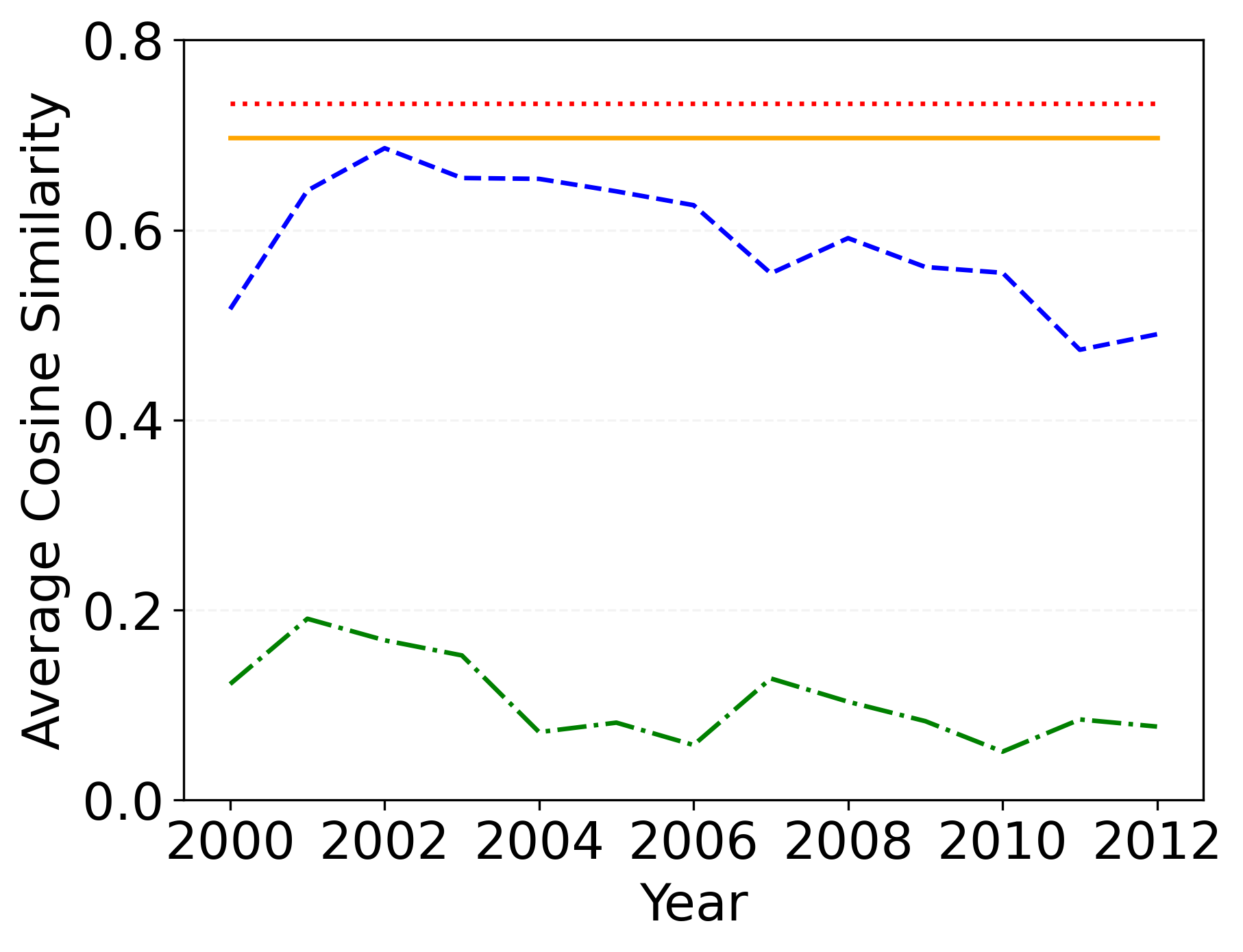} }}%
    \quad
    \hspace{-.4cm}
    \subfloat[\centering Grammar]{{\includegraphics[width=0.41\textwidth]{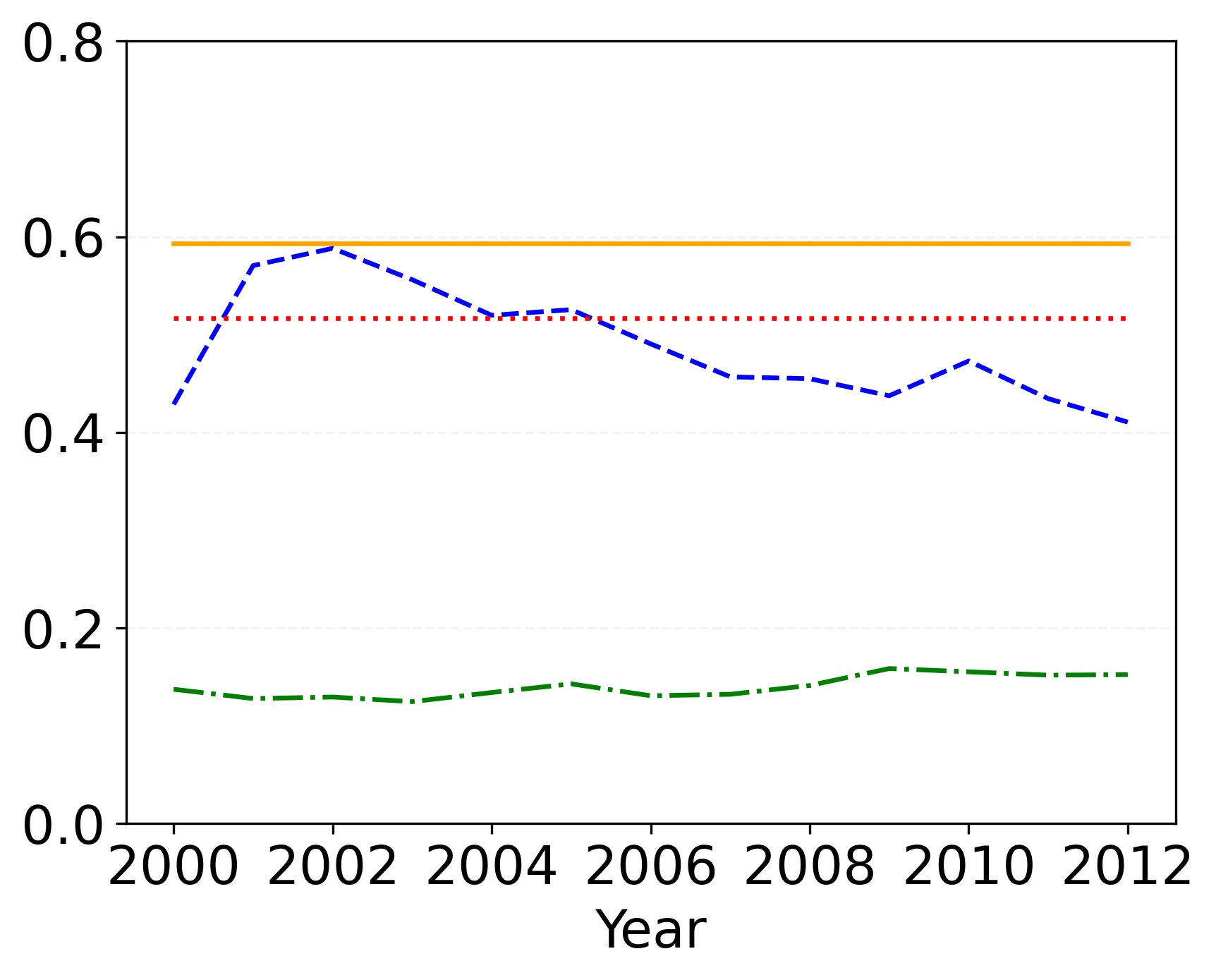} }}%
    \quad
    \subfloat[\centering Geography]{{\includegraphics[width=0.40\textwidth]{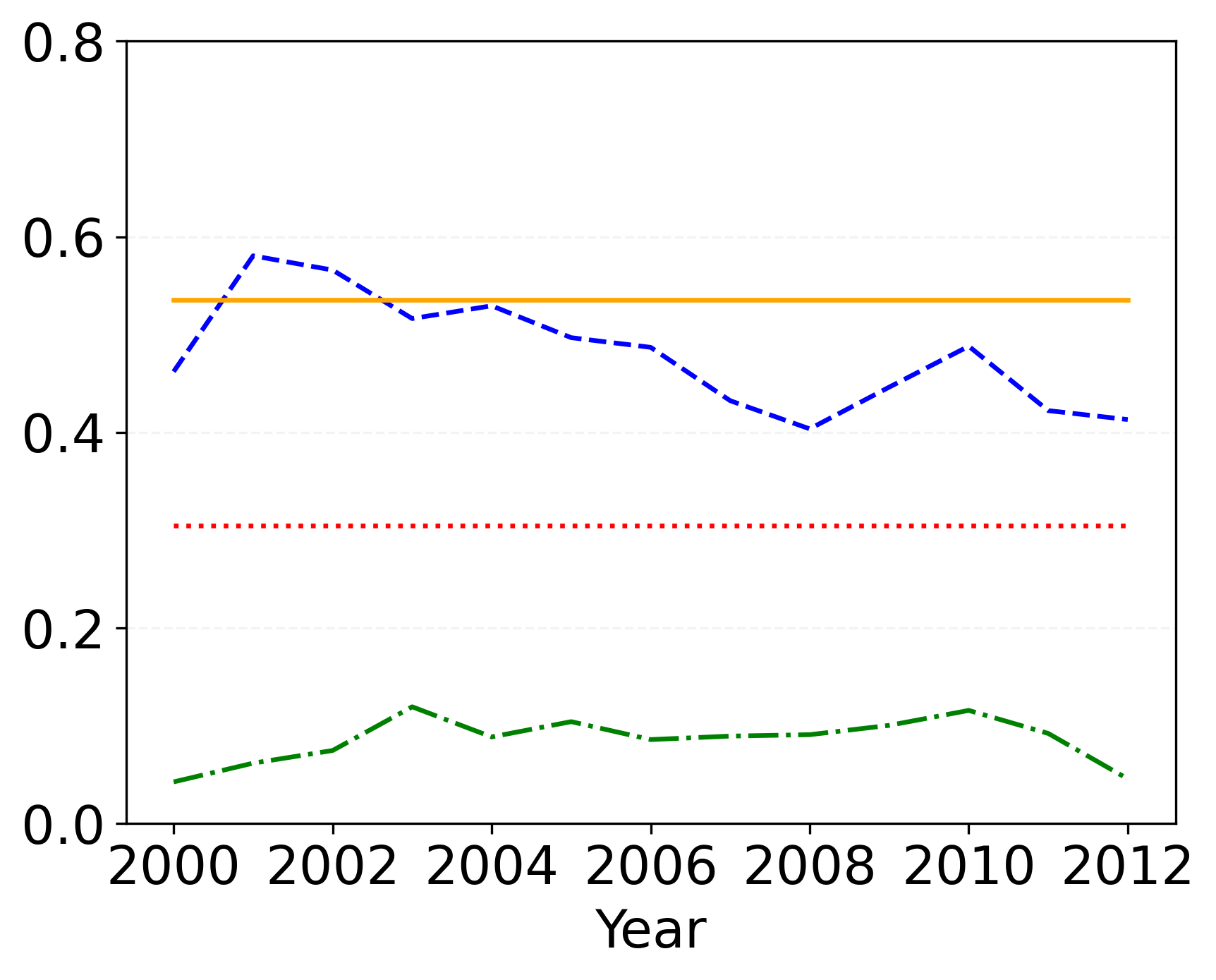} }}%
    \quad
    \subfloat[\centering Currency]{{\includegraphics[width=0.53\textwidth]{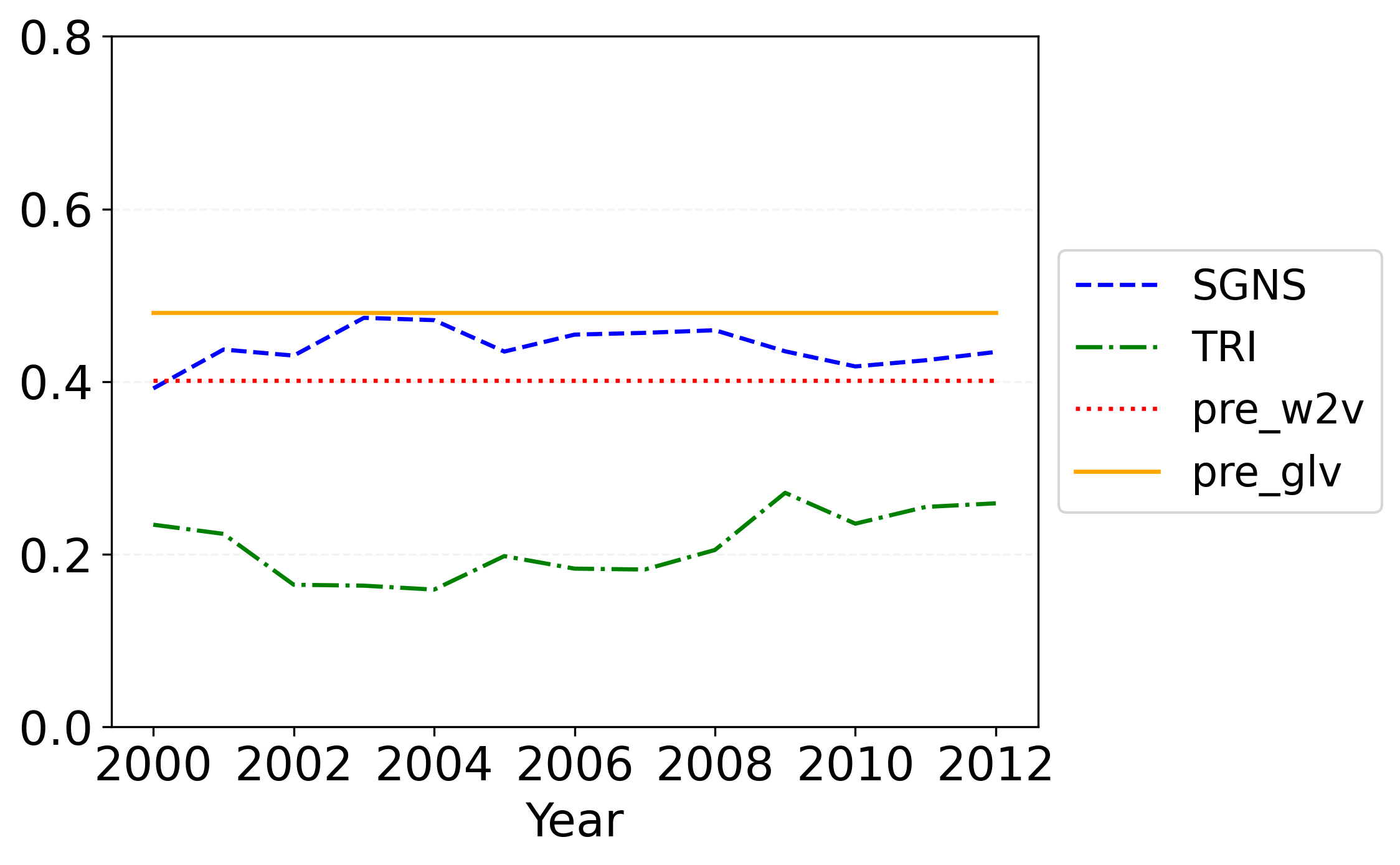} }}
    \caption{Results of the SGNS and TRI embeddings and the two baselines models on the Word Analogy task.}%
    \label{fig:results_analogy}%
\end{figure}

\begin{figure}%
    \centering
    \subfloat[]{{\includegraphics[width=0.41\textwidth]{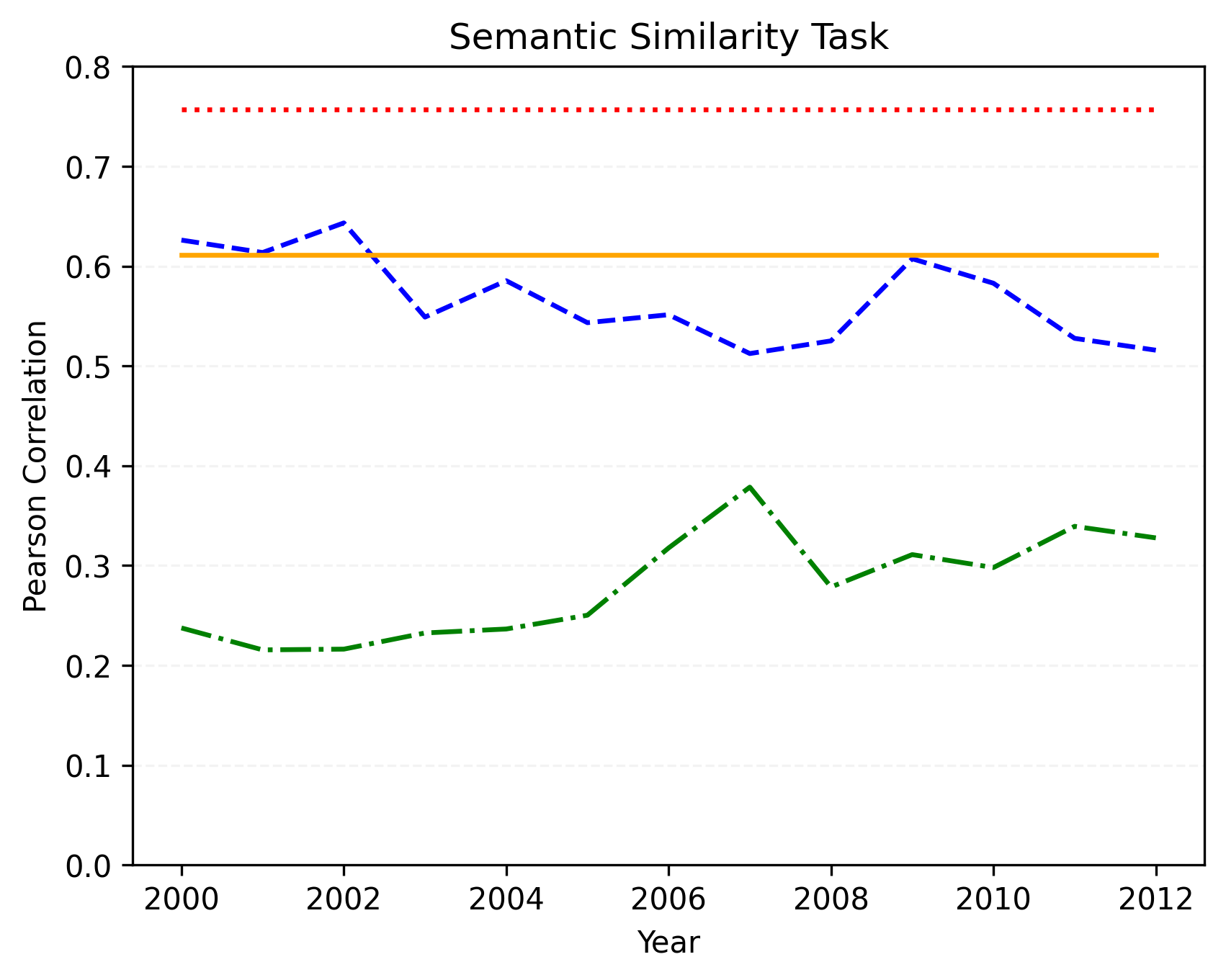} }}%
    \qquad
    \subfloat[]{{\includegraphics[width=0.51\textwidth]{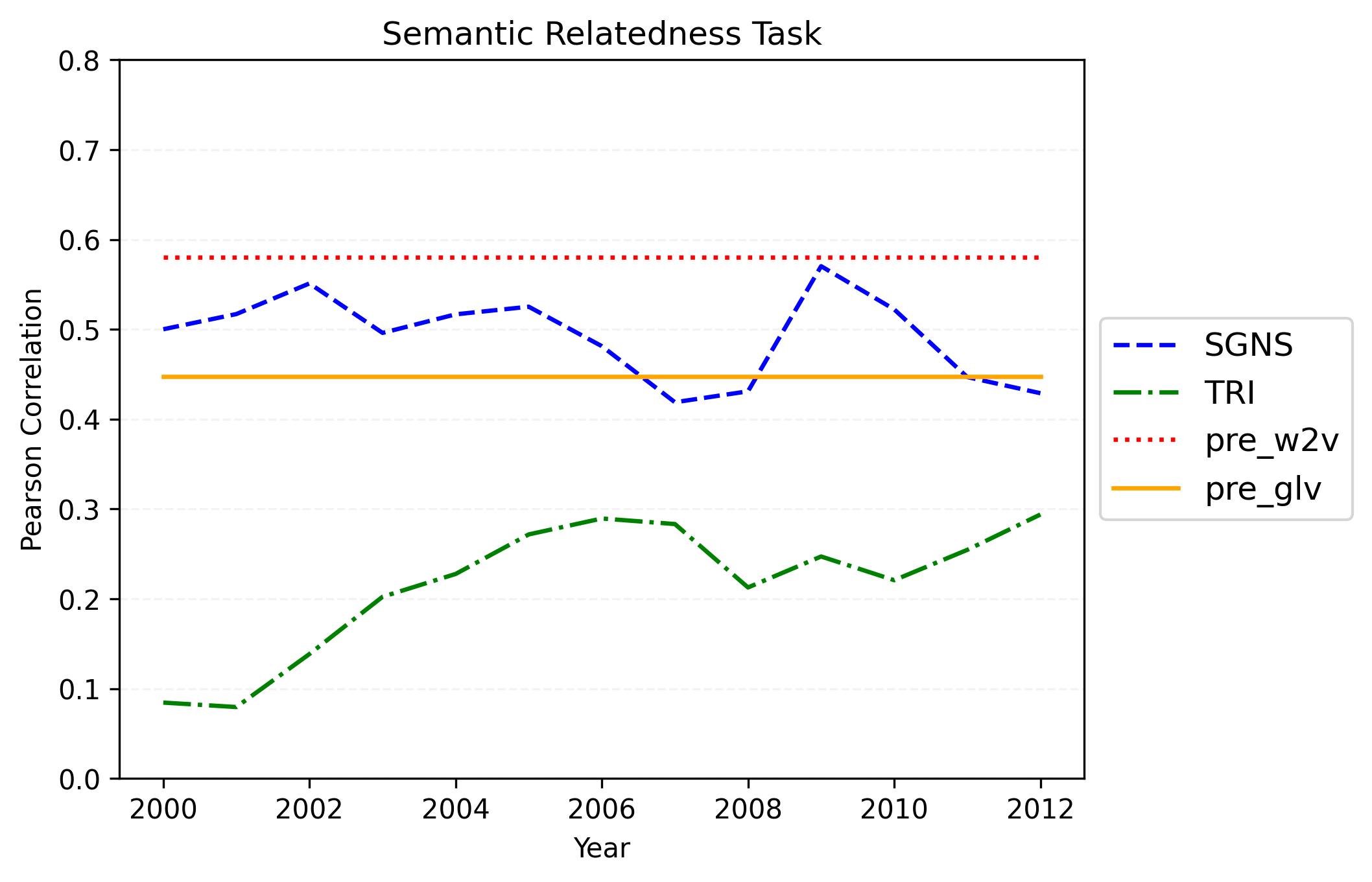} }}
    \caption{Results of the SGNS and TRI embeddings and the two baselines models on the Word Similarity (a) and Word Relatedness (b) tasks.}%
    \label{fig:results_wordsim}%
\end{figure}

\paragraph{Word Analogy} The results are displayed in Figure~\ref{fig:results_analogy}. $pre_{glv}$ outperforms our models in almost all cases. $SGNS$ achieves comparable performance to $pre_{glv}$ and $pre_{w2v}$, especially during the first years; comparably.
$TRI$ performs poorly since it is not able to learn the linear dependency between vectors in the space, which may be due to the \textit{Random Projection} that preserves all distance/similarity measures based on L2-norm, but it distorts the original space and does not preserve the original position of vectors.
$pre_{w2v}$, $pre_{glv}$ and $SGNS$ better capture relationships related to ``Family'' and ``Grammar'' than ``Currency'' in our experiments.

\paragraph{Word Relatedness and Word Similarity} Figure~\ref{fig:results_wordsim} shows the results on the word similarity and word relatedness tasks. Here $pre_{w2v}$ achieves the highest average Pearson correlation score across all years (.76 and .58 for the two tasks, respectively). $SGNS$ performs competitively (.57 vs .61, on average) and outperforms in most cases (49. vs .45, on average) the $pre_{glv}$ baseline for the case of semantic similarity and relatedness, respectively. Importantly, its performance is again consistent across time, ranging from .51 to .64 and from .42 to .57 for the two respective tasks. TRI performs again more poorly and slightly more inconsistently than $SGNS$, with its average (across years) evaluation score ranging from .22 to .38 (average: .28) and from .08 to 0.29 (average: .22). 

The results presented so far show that, overall, the $SGNS$ and $TRI$ embeddings do not outperform the two baselines ($pre_{w2v}$ and $pre_{glv}$) consisting of static word representations in temporally independent tasks. This is partially attributed to the facts that our resources are built on large-scale, yet potentially noisy content and are also restricted primarily to British English, which could impose some geographical and linguistic biases. However, both representations ($SGNS$, $TRI$) can be effectively utilised for a dynamic, i.e. temporally dependent, task which cannot be dealt with static word representations, as we show in the next section.

\subsection{Dynamic Task: Semantic Change Detection}

The major contribution of the $TRI$ and $SGNS$ embeddings of DUKweb consists in their temporal dimension. To exploit this aspect, we measure their ability to capture lexical semantic change over time. 

\paragraph{Experimental Setting} We use as ground truth 65 words that have changed their meaning between 2001 and 2013 according to the Oxford English Dictionary \cite{tsakalidis2019mining}. We define the task in a time sensitive manner: given the word representations in the year 2000 and in the year $X$, our aim is to find the words whose lexical semantics have changed the most. We vary $X$ from 2001 to 2013, so that we get clearer insights on the effectiveness of the word representations. Since our ground truth consists of words that are present in the OED, we also limit our analysis to the 47,834 
words that are present both in the OED and in the TRI/SGNS vocabularies. 

\paragraph{Models} We employ four variants of Orthogonal Procrustes-based methods operating on $SGNS$ from our prior work \cite{tsakalidis2019mining,tsakalidis2020} and two models operating on $TRI$, as follows:

\begin{itemize}
    \item $SGNS_{pr}$ employs the Orthogonal Procrustes (OP) alignment to align the word representations in the year $X$ based on those for the year $2000$;
    \item $SGNS_{pr(a)}$ applies OP in two passes: during the first pass, it selects the most stable (\textit{``anchor''}) words (i.e., those whose aligned representations across years have the highest cosine similarity); then, it learns the alignment between the representations in the years $2000$ and $X$ using solely the anchor words;
    \item $SGNS_{pr(d)}$ applies OP in several passes executed in two rounds: during the first round, it selects the most stable (\textit{``diachronic anchor''}) words across the full time interval; then, it learns the alignment between the representations in the years $2000$ and $X$ using solely the diahcronic anchor words;
    \item $TRI_c$ and $TRI_p$ exploit time-series built by the cumulative and point-wise approach, respectively (see Section \ref{sec:tri})
\end{itemize}

\paragraph{Evaluation} In all of the $SGNS$ models, we rank the words on the basis of the cosine similarity of their representations in the aligned space, such that words in the higher ranks are those whose lexical semantics has changed the most. For the $TRI$-based models, the Mean Shift algorithm \cite{taylor2000change} is used for detecting change points in the time series consisting of the cosine similarity scores between the representations of the same word in each year covered by the corpus. For each detected change point, a p-value is estimated according to the confidence level obtained by the bootstrapping approach proposed in \cite{taylor2000change}, then words are ranked according to the p-value in ascending order.
Finally, we run the evaluation using the recall at 10\% of the size of the dataset as well as the average rank ($\mu_r$, scaled in [0,1]) of the 65 words with altered semantics. 
Higher recall-at-k and lower $\mu_r$ scores indicate a better performance.

\paragraph{Results}
Tables 3 and 4 present the results of our models in terms of $\mu_r$ and recall-at-k, respectively. In both cases, the $SGNS$-based approaches perform better than $TRI$: on average, the best-performing SGNS-based model achieves 27.42 in $\mu_r$ (SGNS$_{pr(a)}$) and 29.59 in recall-at-k ($SGNS_{pr}$). The difference compared to TRI is attributed to their ability to better capture the contextualised representation of the words in our corpus. Nevertheless, TRI has recently achieved state-of-the-art performance on semantic change detection in the Swedish language \cite{schlechtweg-etal-2020-semeval}. Furthermore, despite their superior performance in this task, the Procrustes- and SGNS-based approaches have the shortcoming that they operate on a common vocabulary across different years; thus, words that have appeared at a certain year cannot be detected in these variants -- a drawback is not present in the case of TRI.

Finally, we inspect the semantically altered words that have been detected by each model. Table~\ref{tab:words2} displays the most ``obvious'' and challenging examples of semantic change, as ranked on a per-model basis. It becomes evident that the two different word representations better capture the changes of different words. This  is attributed to the different nature of the two released word representations. Incorporating hybrid approaches operating on multiple embedding models could be an important direction for future work in this task.

\begin{table}[]
\centering
\resizebox{.8\textwidth}{!}{
\begin{tabular}{llllll}
 & \textbf{SGNS$_{pr}$} & \textbf{SGNS$_{pr(a)}$} & \textbf{SGNS$_{pr(d)}$} & \textbf{TRI$_c$} & \textbf{TRI$_p$} \\ \hline
 & cloud                 & sars                     & eris                     & tweet           & root            \\
 & sars                  & fap                      & ds                      & qe              & purple          \\
 & tweet                 & trending                 & follow                     & parmesan        & blackberry      \\
 & trending              & eris                     & blw                       & event           & tweet           \\
 & fap                   & tweet                    & fap                  & sup             & follow          \\ \hline
 & tweeter               & preloading               & unlike                  & status          & eta             \\
 & like                  & chugging                 & chugging               & grime           & prep            \\
 & preloading            & bloatware                & roasting                     & prep            & grime           \\
 & bloatware             & tweeter                  & even                 & trending        & status          \\
 & parmesan              & parmesan                 & parmesan                 & tomahawk        & tomahawk       
\end{tabular}}\caption{Examples of easy-to-predict (top-5) and hard-to-predict (bottom-5) words by our SNGS and TRI models.}\label{tab:words2}
\end{table}

\begin{table}[]
\resizebox{\textwidth}{!}{\begin{tabular}{lrrrrrrrrrrrrr}
Model&2001&2002&2003&2004&2005&2006&2007&2008&2009&2010&2011&2012&2013\\ \hline
SGNS$_{pr}$&36.33&34.03&30.90&26.85&29.29&27.16&26.88&28.60&\textbf{25.16}&27.21&25.69&\textbf{25.83}&29.13\\
SGNS$_{pr(a)}$&36.83&32.30&31.67&27.23&27.27&\textbf{26.31}&\textbf{25.25}&28.15&26.54&\textbf{27.14}&27.59&30.42&28.78\\
SGNS$_{pr(d)}$&\textbf{33.09}&\textbf{28.32}&\textbf{28.91}&\textbf{23.17}&\textbf{25.13}&27.99&30.38&\textbf{25.60}&27.17&28.96&\textbf{25.08}&27.84&\textbf{24.83}\\
TRI$_{c}$&54.65&51.22&56.64&55.63&50.90&55.98&60.96&58.03&58.94&56.59&59.00&45.96&45.72\\
TRI$_{p}$&56.77&59.79&54.81&54.61&53.22&54.39&55.44&55.12&59.76&59.22&53.22&46.30&50.07\\
\end{tabular}}
\caption{Average rank of a semantically shifted word; lower scores indicate a better model.}
\end{table}

\begin{table}[]
\resizebox{\textwidth}{!}{\begin{tabular}{lrrrrrrrrrrrrr}
Model&2001&2002&2003&2004&2005&2006&2007&2008&2009&2010&2011&2012&2013\\ \hline
SGNS$_{pr}$&\textbf{16.92}&23.08&21.54&26.15&\textbf{35.38}&30.77&\textbf{29.23}&32.31&\textbf{38.46}&29.23&32.31&\textbf{36.92}&32.31\\
SGNS$_{pr(a)}$&15.38&\textbf{24.62}&23.08&23.08&30.77&\textbf{35.38}&\textbf{29.23}&32.31&33.85&\textbf{32.31}&33.85&26.15&32.31\\
SGNS$_{pr(d)}$&15.38&18.46&\textbf{27.69}&\textbf{29.23}&\textbf{35.38}&23.08&27.69&\textbf{36.92}&30.77&23.08&\textbf{35.38}&18.46&\textbf{33.85}\\
TRI$_{c}$&7.69&12.31&6.15&3.08&12.31&7.69&7.69&7.69&6.15&7.69&4.62&13.85&12.31\\
TRI$_{p}$&12.31&4.62&7.69&10.77&10.77&4.62&13.85&7.69&3.08&4.62&12.31&21.54&13.85
\end{tabular}}
\caption{Recall at 10\% of our different SGNS and TRI models.}
\end{table}

\section*{Usage Notes}
The DUKweb datasets can be used for various time-independent tasks, as demonstrated in this article. Their major application is for studying how word meaning changes over time (i.e., \textit{semantic change}) in a computational and linguistic context. The instructions on how to run our code for the experiments as well as for further downstream tasks have been made publicly available (see Code Availability).

\section*{Code Availability}

The creation of the described datasets requires several steps, each step is performed by a different software. All the software is freely available, in particular:
\begin{itemize}
    \item the code for the processing of the JISC UK Web Domain Dataset for producing both the WET and tokenized files: \url{https://github.com/alan-turing-institute/UKWebArchive_semantic_change};
    \item the software for building both co-occurrences matrices and TRI: \url{https://github.com/alan-turing-institute/temporal-random-indexing};
    \item the code for the experiments conducted in section 4 can be found at \url{https://github.com/alan-turing-institute/DUKweb} 
\end{itemize}
For information on our input data, refer to: \url{https://data.webarchive.org.uk/opendata/ukwa.ds.2/}

\section*{Acknowledgements}
This work was supported by The Alan Turing Institute under the EPSRC grant EP/N510129/1 and the seed funding grant SF099.

\section*{Author contributions}

A.T. conducted the experiments for the SGNS embeddings and wrote sections 2.4.2, 3.2, 3.3 and 4. 
P.B. conducted the experiments for the TRI embeddings and contributed to sections 1, 2.1, 2.2, 2.3, 2.4.1, 3.1, 3.3, 3.4.
M.B. co-supervised this work, contributed to the design of the experiments and contributed to subsection 2.4.2. 
M.C. co-supervised this work, contributed to the design of the experiments and contributed to subsection 2.4.2. 
B.McG. was the main supervisor of this work, contributed to the design of the experiments and wrote section 1.
All authors reviewed and approved the final manuscript.

\section*{Competing interests}

The authors declare no competing interests.

\end{document}